\documentclass[letterpaper, 10 pt, conference]{ieeeconf}  

\usepackage{cite}
\usepackage{graphicx}
\usepackage{multicol}
\usepackage{caption}
\usepackage{url}
\usepackage{booktabs,chemformula}
\title{\LARGE \bf
Enhanced Transfer Learning for Autonomous Driving with \\ Systematic Accident Simulation}

\author{Shivam Akhauri\thanks{e-mail: sakhauri@umd.edu}  %
     \and Laura Y. Zheng\thanks{e-mail: lzheng98@umd.edu} \\ %
       University of Maryland at College Park \\ %
       Supplementary material at \url{https://gamma.umd.edu/etladsas}
    \and Ming C. Lin\thanks{e-mail: lin@cs.umd.edu}} %

\begin{document}

\maketitle
\thispagestyle{empty}
\pagestyle{empty}


\begin{figure*}[hbt!]
\begin{center}
\includegraphics[width=7in]{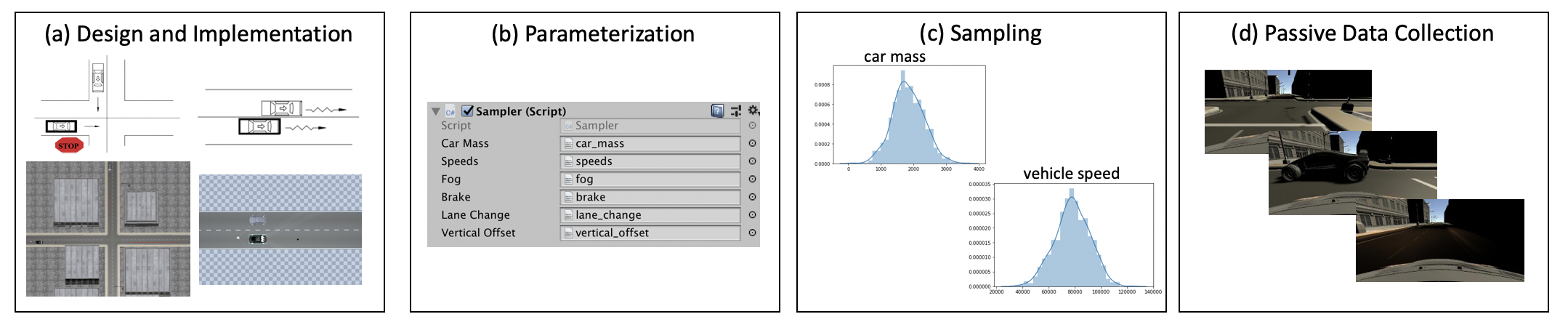}
\end{center}
  \caption{Overview of systemic scenario generation, detailed in section  \ref{section-3}. The generation can be described in four parts: (a) the design and implementation of pre-defined scenarios from the NHTSA pre-crash scenario descriptions \cite{NHTSA}, (b) the generalized parameterization of those scenarios based on physics and visual-related characteristics, (c) sampling of those parameters based on statistical data and reasonable assumptions, and (d) passive data collection, which automates previous parts to collect data from scenarios sampled with different parameter values. Since the goal is to generate accident data, each simulation runs for 10 seconds in order to enable both vehicle agents sufficient time to collide or pass each other.}
  \label{fig-1}
\end{figure*}

\begin{abstract}
Simulation data can be utilized to extend real-world driving data in order to cover edge cases, such as vehicle accidents. The importance of handling edge cases can be observed in the high societal costs in handling car accidents, as well as potential dangers to human drivers. In order to cover a wide and diverse range of all edge cases, we systemically parameterize and simulate the most common accident scenarios. By applying this data to autonomous driving models, we show that transfer learning on simulated data sets provide better generalization and collision avoidance, as compared to random initialization methods. Our results illustrate that information from a model trained on simulated data can be inferred to a model trained on real-world data, indicating the potential influence of simulation data in real world models and advancements in handling of anomalous driving scenarios.

\end{abstract}



\section{INTRODUCTION}
    Transfer learning has potential to infer knowledge from the simulation domain to real-world scenarios. Transfer learning excels where data from different feature spaces or domains are used in conjunction to train a model. Often machine learning applications lack an adequate amount of real-world data. Insufficiency of data can be compensated by generating simulated data in the virtual world and using transfer learning as a means to bridge that gap. 
    
    High-quality, annotated real-world driving data is abundant. The surge in data availability has revolutionized driving models' abilities to learn proper driving behavior. However, this marginalizes edge case data in the scope of learning. Current state of the art machine learning models in autonomous driving are limited by the safe-driving nature of real-world data. When a vehicle is equipped with dozens of sensors for data collection, it is impractical and unrealistic to capture a proportional amount of accident or improper driving data. However, unsafe and accidental scenarios cannot be ignored. While accidents among autonomous vehicles and even among human drivers remain relatively rare, when compared to proper accident-free driving, the high societal costs and safety risks associated with a single accident make it an urgent problem to address. 
    
    Research in autonomous driving can be categorized into different areas: perception, planning, and control. 
    Our approach encompasses all three through an augmented dataset. Instead of the use of all sensors, we focus on image data from dash-camera views. 
    We explore the possibility of using simulated virtual accidents to complement existing datasets and the possibility that behaviors learned in the virtual domain are transferable to the real world. We hypothesize that using systematically-generated accident data to complement real-world datasets through transfer learning can improve autonomous driving models' generalization to diverse driving scenarios. 
    
    Recent state of the art reinforcement learning algorithms suffer from a common problem: the training data that is generated is dependent on the current policy because the agent is generating its own training data by interacting with the environment. This leads to interaction with new scenarios that the agent might not have encountered previously. These scenarios in autonomous car setting could include collision-inducing situations. If the learning rate is too high, the policy update pushes the policy network to the parameter space where the network collects the next batch of data under a very poor policy, causing the model to become unrecoverable. Since an inverse reinforcement learning model intuitively involves the reinforcement learning algorithm, the issue is implied.
    
    Another problem with proximal policy optimization (PPO) and other policy gradient approaches is that they are less sample-efficient, because they only use the collected experience once for performing updates. 
   The PPO method's loss function depends on the log of probabilities from the output of the policy network and estimate of the relative value of the selected action in the current state. Both of these components that construct the loss of the policy gradient approach are composed of a sequential neural network architecture made of convolutional neural layers similar to that of the imitation learning pipeline. 
    
    While there is a perception that there is an abundance of data from dash cam footage widely available online (e.g. YouTube), the data is not usable for several reasons: (1) The image data is often too low-resolution, (2) the camera angle is fixed and not sufficient enough to reconstruct the whole scenario or account for moving variables outside of the frame, and (3) the image data is often not labeled; our experiments required a steering angle label. Thus, we constructed our own basic simulation environment in order to account for lack of control in existing crash data.
    
    We propose that both of these challenges of the policy gradient approach can be addressed by inducing transfer learning to the individual components of the policy gradient approach. We prove our hypothesis by utilizing imitation learning architecture that comprises of sequential neural network design, similar to that of the advantage-calculating neural networks of the PPO method. We take a baseline imitation learning model without transfer learning and our model that exhibits transfer learning, then compare their performance to demonstrate that transfer learned models are able to handle new scenarios much better. Inducing a first-level training of the model in a simulated world allows the agent to explore and generate collision leading scenarios. We further cover more unseen scenarios by using data from systemically-generated accident data that incorporates a wide distribution of possible accident scenarios. We use existing literature on traffic accidents \cite{NHTSA} to recreate common accident scenarios to the best of our ability. Although the scenarios are simple, they capture the elements from literature that lead to the accident. As future work, visual enhancements and moving objects can be incorporated into the simulation as well.

    
    In summary, the contributions of this paper include 
    \begin{enumerate}
        \item A systematic approach to generate accident data from pre-crash scenarios as defined by the NHTSA report~\cite{NHTSA}.
        \item A proof of concept in combining simulated driving data and real-world driving data to train an imitation learning model.
        \item Improvements in steering wheel predictions from the-state-of-the-art end-to-end learning models to prove its efficiency and to extend it on policy gradient approaches \cite{baseline}.
    \end{enumerate}

\section{Related Works}
    \subsection{Simulated Driving Data}
        Related studies have explored the use of realistic driving simulations to provide a cost-efficient and controlled alternative to real world data. These simulations utilize the power of game engines, which provide for realistic image generations and built-in physics. Examples include CARLA, Drive Constellation by NVIDIA, and AirSim from numerous industry leaders in autonomous driving \cite{CARLA, AirSim}. Some studies sought to expand upon the versatility of these systems by boosting the realism of the images generated by rendering 3D models into real-world background images, such as the Augmented Autonomous Driving Simulation model \cite{AADS}. In parallel, our accident scenario generation relies on video game engines for built-in physics and object models. However, instead of representing an all-safe driving environment as close as possible to reality, our simulation focuses on covering as much of the parameter space as possible for selected scenarios. 
        
        Aside from driving simulators meant for research, video games have also been extensively used as a source of data. Grand Theft Auto, an open-world video game involving extensive urban area navigation and user-controlled driving, has been used in reinforcement learning research for autonomous driving \cite{GTA}. While video games such as Grand Theft Auto provide a visually rich environment, which include pedestrian models and open world movement, there is little to control agent-based accident scenario generation. Our simulator aims to control accident scenario generation over a continuous distribution, holding all other factors constant. 
        
        There are many existing driving simulators backed by industry initiatives, as well as equipped with realistic graphics and diverse environments. However, these simulations lack the flexibility to design and control the systemic generation of accident scenarios. We focus on simulation data generation, where we systemically generate accident scenarios by parameterizing common pre-crash scenarios and sampling those parameters over a distribution. 
        
\begin{figure*}[hbt!]
\begin{center}
\includegraphics[width=6in]{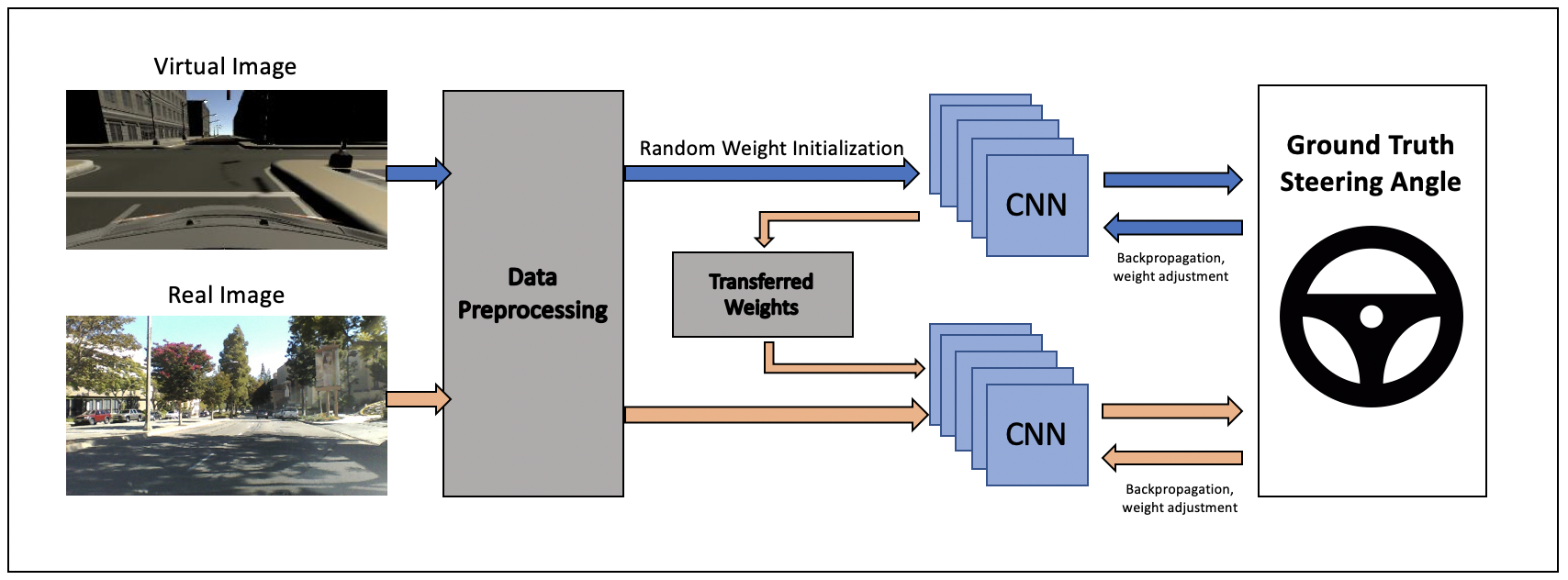}
\end{center}
  \caption{Transfer learning experiment design. Our pipeline includes two stages: Stage 1, represented by blue arrows, involves training of the model using the simulation data that incorporates accident and collision-inducing scenarios. Stage 2, represented by orange arrows, involves training the model on real-world dataset with its neural network weights initialized with weights obtained from stage 1. Stage 2 only begins training once stage 1 is finished training. This leads to better generalization and more effective collision avoidance, as compared to the baseline model trained with transfer learned weights \cite{baseline}. }
  \label{figure-2}
\end{figure*}

    \subsection{Transfer Learning}
        Transfer learning can improve learning by transferring information from a related domain. For example, having knowledge in mathematics and statistics can help one understand concepts in machine learning better. In the last decade, transfer learning research has worked towards adapting to different domains, applications, and methods. Karl Weiss' "A Survey of Transfer Learning" formally defines transfer learning, presents current state of the art, and reviews applications applied to transfer learning \cite{TransferLearningSurvey}. Specifically, we explore the use of transfer learning in autonomous driving, with a focus on transfer from the virtual world to the real world domain. 
        
        In one work, the same circumstances for virtual-to-real transfer learning were applied to wilderness traversal \cite{VirtualRealWilderness}. Due to a lack of real-world wilderness data, cost-efficient virtual data was utilized in virtual-to-real transfer learning. However, one major difference is that a small subset of real-world data was available for validation. In our case, not even one sample of annotated real-world data was available for ground-truth validation of our learned model. In applications towards autonomous driving, transfer learning has been popular in reinforcement learning models. One such paper presented the first successful virtual-to-real driving policy with reinforcement learning \cite{VirtualToRealRL}. Since reinforcement learning models are difficult to train in the real-world, information learned from virtual driving data was used to train a driving policy. While collecting data in the real world is just as infeasible, our experiments aim not only to train a proper driving policy, but also inherent hazard recognition and avoidance. Some sought to solve the domain discrepancy in the other direction; for example, Yang et al. presented Real-to-Virtual domain unification, in order to simplify real world data to its much simpler virtual counterpart \cite{UnsupervisedDomainUnification}. 
        
        Imitation learning for autonomous driving has been successful in cases without extensive sensor measurements. In Bojarski's "End to End Learning for Self-Driving Cars", a convolutional neural network (CNN) was trained to recognize steering angles off of image data \cite{baseline}. 
        We draw inspiration from Bojarski et al. by implementing their end-to-end system, while also attaching our transfer learning and complementing accident scenario data. The implementation by Bojarski et al. is also appealing due to the sole use of image inputs. 
        
        Imitation learning addresses two main problems presented by current state-of-the-art reinforcement learning algorithms: (1) training data generated is dependent on the current policy and (2) sample inefficiency due to one time use of the collected experience for backpropagation. For the first issue, the policy update may push the network towards a parameter space to become unrecoverable. The second issue presents a challenge of underfitting on diverse scenarios in the environment for model training . 
        
        The proximal policy optimization (PPO) loss function depends on the output of the policy network, and estimates the value of an action in the current state. The components that construct the loss of the PPO approach are composed of a sequential neural network architecture that is similar to that of an imitation learning pipeline. In other words, using an imitation learning model directly addresses these two challenges presented to current state-of-the-art reinforcement learning models. 
        
        To show effects of transfer learning on accident data, we implement the end-to-end system by Bojarski et al. \cite{baseline} with an imitation learning model, and consider that as our baseline model. In conjunction, we utilize imitation learning for our model in order to take advantage of what has already been done. 
        We show that creating a two-level training process, as opposed to single-level, with the augmented dataset better generalizes safe driving and avoids collisions otherwise unavoidable with the baseline method. In this way, we show that simulated accident data positively improves the results using only real-world driving data. 

\section{Methodology}
    \label{section-3}
        We based scenario design on those defined in the vehicle-on-vehicle accident report by the National Highway Transportation Safety Administration \cite{NHTSA}. While it's difficult to simulate all possible random accident scenarios in the universe, we considered this report to be a starting point where common scenarios are defined. 
        
        First we set up two types of road layouts common to many scenarios: highway and intersection. The highway consisted of one straight road, with two lanes. Depending on the scenario, the two lanes may be going the same direction or in opposite directions. The intersection environment consisted of a four-way urban intersection, with traffic lights or four-way stop signs. 
        
        These two environments are created with the statistics from the pre-crash report in mind \cite{NHTSA}. Scenarios will have statistics on what conditions crashes occurred in. For example, the NHTSA reports that 94\% of "running red light" accidents occur on "intersection or intersection related at 3-color traffic signal", while 70\% of "opposite direction" accidents occur on "non-junction without traffic controls" environments. For some scenarios, the occurrences between intersection vs. non-intersection environments are much closer, so environment was determined based on the diagram provided by the report.
        A list of these scenarios can be found in Table~\ref{tb-3} below.
        
        \begin{table}[h]
          \centering
          \begin{tabular}{@{}cc@{}}
            \toprule
            Scenario Description & Environment Type\\
            \midrule
            Running Red Light & I \\
            Running Stop Sign & I  \\
            Turning/Same Direction & I  \\
            Changing Lanes/Same Direction & H  \\
            Drifting/Same Direction & H \\
            Opposite Direction/Maneuver & H \\
            Opposite Direction/No Maneuver & H \\
            Rear-End/Lead Vehicle Accelerating & H \\
            Rear-End/Lead Vehicle Moving Slower & H \\
            Rear-End/Lead Vehicle Decelerating & H \\
            LTAP/OD at Signal & I \\
            Turn Right at Signal & I \\
            LTAP/OD at Non-Signal & I \\
            Straight Crossing Path at Non-Signal & I \\
            Turn at Non-Signal & I \\
            \bottomrule
          \end{tabular}
          \caption{The scenarios modeled as described by section \ref{section-3}. Nearly every scenario in the NHTSA \cite{NHTSA} pre-crash scenarios report is visually modeled as an intersection (I) scenario or a highway (H) scenario. Wassim et al. define 37 pre-crash scenarios and give statistical data on 17 of those 37 scenarios. Five of the remaining scenarios involve rear-end collisions, which were modeled, but not used in the experiment, as our training data involved only front dash-board images.}
          \label{tb-3}
        \end{table}
        
    \subsection{Parameterization}
        We define six generalized parameters that cover every scenario: car mass, speed, fog, brake force, lane change distance, and vertical offset. These parameters were determined based off of pre-crash scenario descriptions from the NHTSA report \cite{NHTSA}. For example, atmospheric conditions was an environmental factor leading up to crash scenarios, so we reflect that in our simulator with fog. Some parameters were difficult to model accurately in Unity, such as slippery road conditions. "Slippery" can cover a wide range of friction, from light rain to black ice. While there are capabilities in Unity to adjust the friction of object textures, very little was readily available for adjusting friction on specifically a road asphalt texture.
        
        We define lane change distance as the horizontal distance, or distance parallel to the road, a car travels when completing a lane change. Conversely, the vertical distance is defined as the distance a car shifts when changing from one lane to the next. These two parameters are common to all highway scenarios, but not used in intersection scenarios where cars are not changing lanes. In addition, these two parameters directly affect the waypoint paths of vehicle agents in the accident simulations. Other parameters affect aspects of the car and of the environment.

    \subsection{Sampling}
        We used a case study by the U.S. Department of Transportation to justify the speed distribution \cite{SpeedLimits}. For car mass, we referenced a study on fuel economy standards that provided statistical data on masses of different vehicles. Specifically, we used the category for “all cars”, since the NHTSA report involved vehicle-on-vehicle collisions \cite{CarMass}. For parameters with real-world measurements, we sampled based on a Gaussian distribution to most accurately reflect scenarios in the real world. The number of simulations we can generate per scenario is unlimited, so we chose not to use systematic sampling. In data collection, the interval at which to sample may become unclear. To keep assumptions as minimal as possible, we used simple random sampling from the distributions we generated, since we want every value in the distribution to have an equal probability to be selected. 
        
        Some parameters were considered arbitrary due to the lack of statistical reports or by nature. Visibility was modeled in Unity with a fog, in order to reflect different weather conditions. We also sampled this parameter from a half-normal distribution for clearer visibility. Fog was tuned arbitrarily by hand due to Unity's implementation of fog; for values exceeding 5\% visibility, the dash view becomes obscure even for the human eye. Thus, we determined to use a half-normal distribution, where most cases would provide visible data, while others may have a slight fog. Little measured data on brake force, lane change distance, and vertical offset exist. Thus, we sample these arbitrarily on a normal distribution where averages are centered around typical driving behavior. For example, the average parameter values for vertical offset would be the distance for the vehicle to travel from the center of one lane to the center of the next. 
        
        While it may seem more intuitive that non-normal distributions should be considered when generating accident scenarios, we would also like to point out that each vehicle agent is set up to be an equal distance from the area of contact. In other words, vehicles are placed in the same distance from the center of each intersection, which each vehicle has the objective of crossing the intersection. Thus, in order to generate more driving accidents, we want to model average driving behavior in situations where vehicles are interacting in close proximity. 
      

\begin{figure*}
\begin{center}
\includegraphics[width=4in]{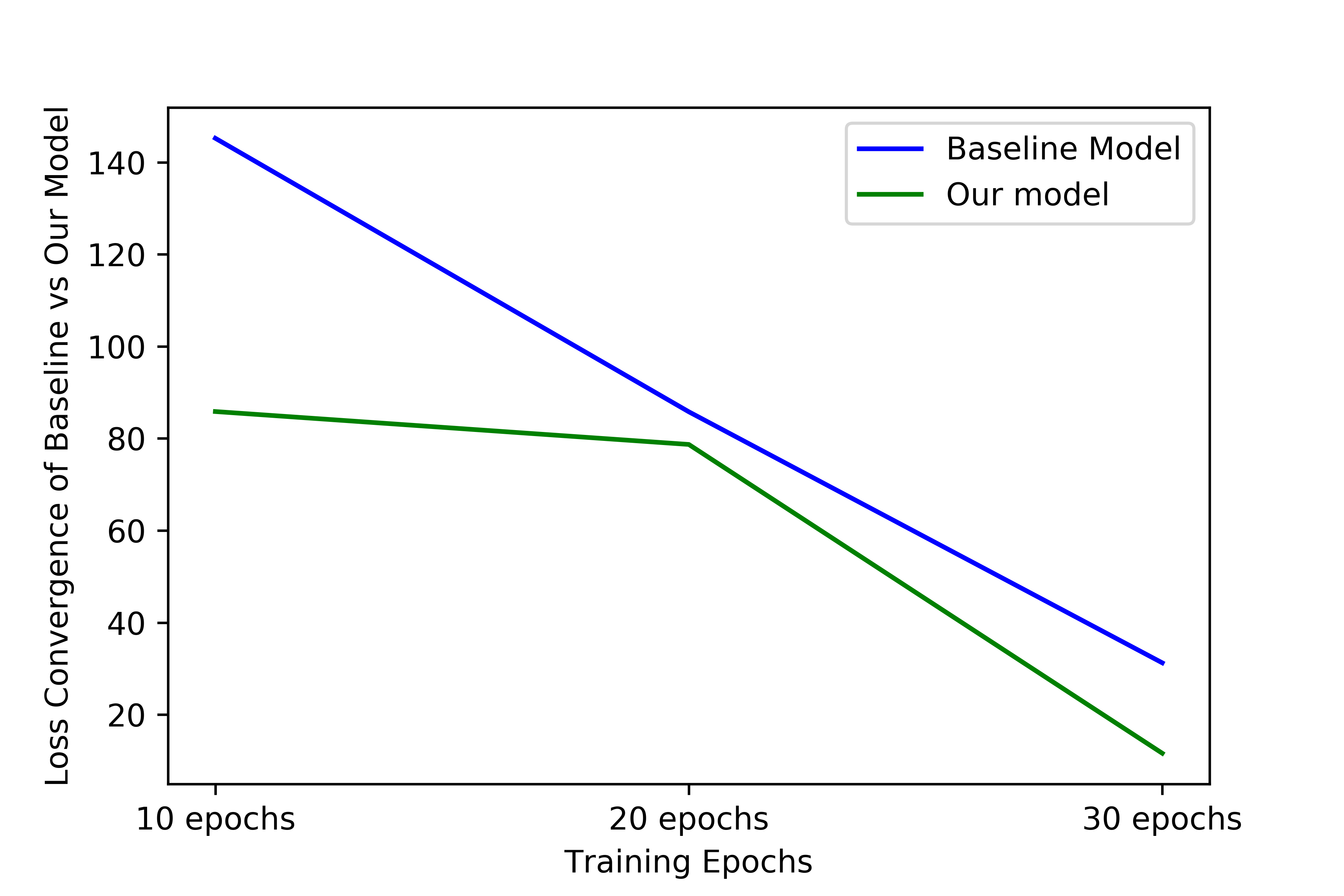}
    \caption{Comparison of validation loss and accuracy between baseline model and our complemented transfer learning model.We see that the loss convergence for the complemented transfer learning model is much better than the baseline model at various stages of the training. We also see better training accuracy of the transfer learning model ranging from 5\% after 10 epochs to 9\% after the 30th epoch .}
    \label{table-2}
\end{center}
\end{figure*}
 
\begin{figure*}
\begin{center}
\includegraphics[width=6.5in]{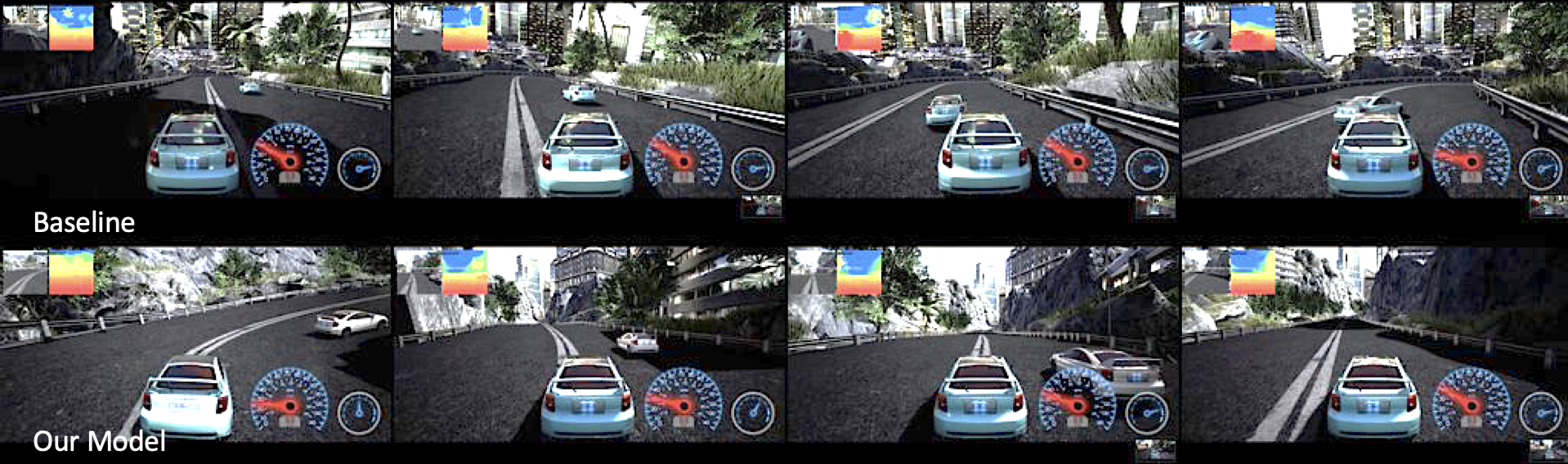}
    \caption{Qualitative results in transfer learning to DeepDrive simulator on a non-intersection scenario. The top row represents a progression of images in lane change using the baseline method with randomized Xavier initialization. As the car is changing lanes, it collides with the car in front. The top right image shows this collision. By comparison, the bottom row represents the same action with the complemented transfer learning method, but avoids collision with the other vehicle by passing. This example shows a non-intersection scenario where the addition of accident scenario data improves safety of autonomous driving.}
    
    \label{fig-3}
\end{center}
\end{figure*}

\begin{figure*}
\begin{center}
\includegraphics[width=6.5in]{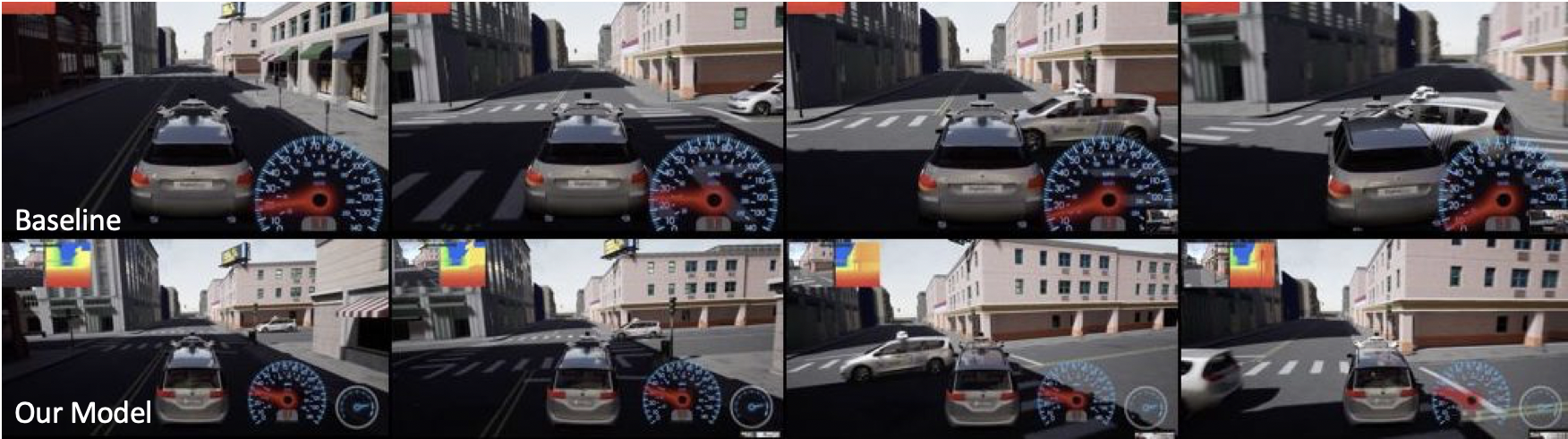}
    \caption{Qualitative results in transfer learning to DeepDrive simulator in an intersection scenario. Likewise from Figure~\ref{fig-3}, the top row represents a progression of images in lane change using the baseline method with randomized Xavier initialization. In the top right image, the collision occurs with the vehicle attempting to wrongfully cross the intersection. In comparison, the bottom row represents the same scenario with the complemented transfer learning method, but avoids collision with the other vehicle by turning to the right. This example illustrates an intersection environment where the addition of accident scenario data improves safety of autonomous driving.}
    
    \label{fig-4}
\end{center}
\end{figure*}

    \subsection{Data Collection}
        
        We model each vehicle in an agent-based fashion. In other words, each vehicle is its own independent agent, with no external factors affecting its actions or dependence on other vehicles. For each scenario, two vehicles are involved. One vehicle is defined to be “properly driving”, while the other is defined as “dangerously driving”. The image data is collected from the properly-driving car, since our goal is to train the model to react to hazards in the environment. For intersection scenarios, vehicles are placed on different sides of the intersection at an equal distance from the center of the intersection. Wheel angle, braking, and acceleration is determined by a waypoint system. All car capabilities are modeled by the car from Unity’s standard assets. Once each vehicle reaches the end of its waypoint segment, braking begins and acceleration becomes zero. Waypoints are considered more as guidelines for the path of the vehicle, and only affect the steering angle during runtime; the vehicle does not follow the waypoint segment strictly. Each vehicle’s mass, speed, and characteristics of its waypoint segment are randomly sampled from the distribution at the beginning of runtime. 
        
        The scenarios are set up in 10-second time intervals, since the point at which vehicles cross paths can be very brief. A longer time interval would be arbitrary, as it would only reflect proper driving found pervasively in real world datasets. Some runs may not produce a collision, but may be very close to collision. These examples are just as important as scenarios with physical contact. Safe-driving humans avoid driving too close to other cars, even if the dangerously-driving vehicle does not make contact in the result. From about 1000 total runs, the rate of scenarios with physical contact between vehicles is 28.97\%, which is significant in comparison with the sparsity of accident data from real-world datasets. 
      

    \subsection{Transfer Learning}
        With transfer learning techniques for our predictive modeling, we aim to accelerate the training and improve the performance of the learning model in steering angle prediction of real-world images. We achieve this through the transfer of model weights from driving scenarios in the virtual (simulated) domain to the model trained with the real-world datasets.
        
        Our model architecture consists of five convolutional layers and four fully connected layers. During training with virtual data, initialize each of these weights with standard Xavier initialization. These uniformly random distributed weights are tuned with backpropagation, when the neural architecture trains with the simulation data. After training, the weights are tuned to provide appropriate steering predictions in the simulated environment for diverse driving scenarios.
        
        Once the model is trained on virtual accident data, which includes the data from accident scenarios, reaches completion, we train the same neural architecture with real-world datasets. Instead of initializing the weights of this model with Xavier initialization or with other uniformly random distributed weights, we initialize them with tuned weights from the predictive model trained on the simulated datasets, until results reach a convergence.


\begin{figure*}
\begin{center}
\includegraphics[width = 5in]{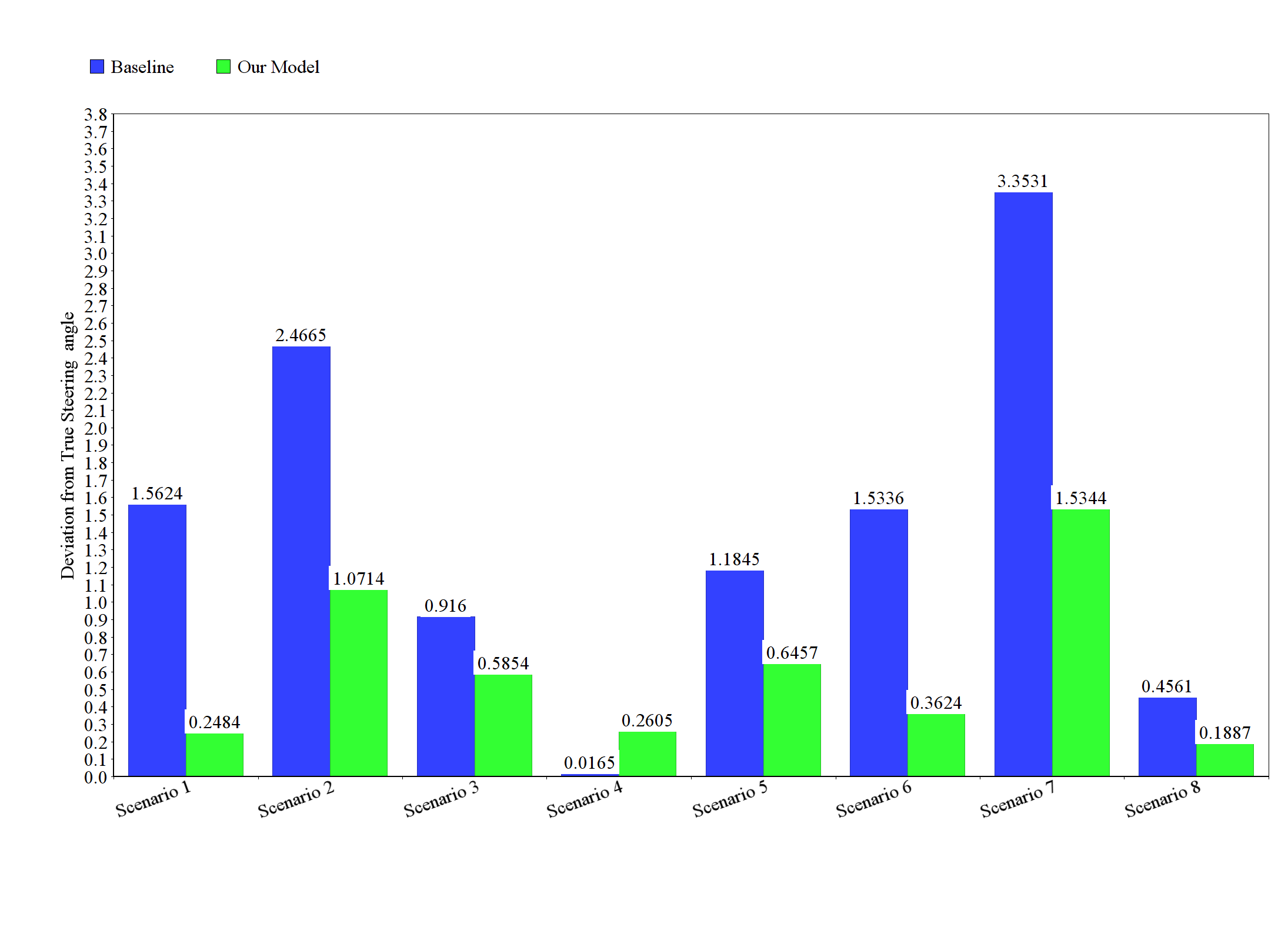}
    \vspace*{-3em}
    \caption{Comparison between baseline \cite{baseline} model and our model that utilizes simulated accident data, in predicting the steering angle for real world images. The deviation, measured in degrees, from the true label steering angle is plotted for eight sample images. The baseline model generally {\em deviates more} than our method, suggesting that complementing models with simulated accident data through transfer learning reflects driving closer to that by real human drivers.}
    \label{fig-5}
\end{center}
\end{figure*}

\section{Results}
 
\subsection{Qualitative Simulation Results}
    Because the output of the systemic accident scenario simulations is image data, the results are presented qualitatively. Sample images from the data simulator are shown in the appendix \ref{appendix-1}, from both intersection and non-intersection environments.

\subsection{Comparison in Convergence Time}
    To validate that our learning model appropriately inferred road features and collision leading scenarios better, we compared the baseline model initialized on a uniformly random distribution with our transfer learning model. For this, we maintained training hyper-parameter paradigms and training sets.
    
    We observed much faster training convergence on the transfer learned model, with the validation loss reaching 85.85 at the end of 10 epochs. Compared with the baseline, which reached a comparable loss at the end of 20 epochs, the time to convergence is significantly less. On average, the validation accuracy was 8\% better on the transfer learned model on the same test set. We saw the validation accuracy improve even further for lower learning rates. These results validate that the important features learnt by the model from the simulator data  facilitates the training convergence when training the CNN on the real-world dataset. Details on the comparison of convergence time are in \ref{table-2}. For this set of results, the validation loss was 0.35.
    
    The test results in Table \ref{table-2} were obtained by keeping the same batch size, while training the CNN on the same real-world dataset. We observe that the results emphasize the faster convergence and better validation accuracy for the model trained on the real world data set with weights initialized from another CNN model trained on the virtual world images. For this set of results, the validation loss was 0.35.

\subsection{Validation in Real World}
    In order to demonstrate the quantitative effect of the pre-trained weight model on the transfer learning model, we further improved upon the convolutional network with regards to loss and accuracy. On this improved CNN, we obtained a validation loss of 0.28 and a +9.8\% accuracy difference for the transfer learning model, compared to the baseline. More information about the results can be found in Table \ref{table-2}. This also further suggests that a better quality model trained solely on virtual data has considerable impact in a transfer learned model trained on real world data.
    
    To demonstrate the impact of a better-trained model on steering prediction, we compare trajectory predictions on single-view images against that of the baseline. Using the real world data set, we let both models predict the steering angle for the test dashboard images and compare those predictions with the true steering angle which the expert, human driver exercised while driving during the data set collection. 
    From a total of 166 test images from the real world, we measured the average deviation for both models from the true steering angle. On the baseline inspired model by Bojarski \cite{baseline}, the average deviation from ground truth is 3.3200, while the average deviation on the transfer learning model is 2.2803, showing an approximate 31.31\% improvement in steering wheel prediction. Qualitative results showing the affect of accident scenario data can be observed in  Figure~\ref{fig-3} and \ref{fig-4}. 
    
    The steering wheel results suggest that high-level knowledge from the virtual world data can be inferred and utilized in real-world scenarios, including that of accident avoidance behavior. 
    


\section{Conclusion}

In this paper, we show that transfer learning using simulated accident data, with automatic sampling and parameterization, leads to better generalization to more diverse scenarios. As stated earlier, much of the individual components of the policy gradients reinforcement learning algorithms involve convolutional neural networks, where the transfer learning approach could be leveraged to facilitate better convergence and sampling efficiency.

\subsection{Limitations}
Our current in-house accident simulator can further benefit from a more diverse, dynamic, and realistic vehicle and environment models. This has implications for training with image data, as the environment remains static and likely contributes to bias in the learning process. Had the accident scenarios taken place with different visual environments, the results may be improved. 
\par One limitation is obtaining annotated real world crash scenarios, along with proving the ability to detect possible accidents. Although accident data was incorporated into the training of the convolutional model for pre-trained weights, the bias in data and lack of a labeled real world validation set made it difficult to directly test the effectiveness of the data set. 

\subsection{Future Work}

    The framework for the accident scenarios is not limited to the in-house simulator in this project. The same scenarios can be replicated in simulators with varying environments and higher-quality visualization, in order to achieve images closer to the real-world domain. This could improve the diversity in environment scenes. With more visual variation in the environment.  
    
    Furthermore, addressing this potential would enable for other methods of domain transfer. With more diverse image data, it would be a feasible hypothesis that current supervised and unsupervised methods of image domain transfer would be successful, considering that some studies were able to convert virtual driving images to the real world domain \cite{imgtransfer1, CycleGAN,UnsupervisedImgTranslation}. This, combined with state-of-the-art road image segmentation, can be combined as another method for domain transfer \cite{RoadSegmentationLSTM,Long2015FullyCN}.
    
    \section*{Acknowledgments} This work is supported in part by the Army Research Office and
    Elizabeth Stevinson Iribe Professorship. \\ \mbox{}


\bibliographystyle{plain}
\bibliography{root}


\begin{appendix}
    \section{Sample Accident Simulation Images} 
    \label{appendix-1}
        \subsection{Intersection - Running Red Light}
            \includegraphics[width=3.2in]{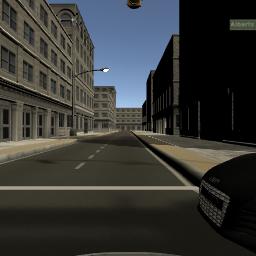}
        \subsection{Non-intersection - Lane Change / Same Direction}
            \includegraphics[width=3.2in]{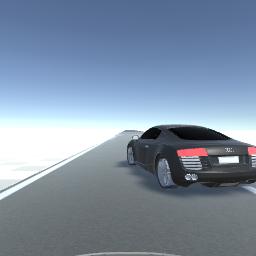}
    
    \section{Example of a Pre-crash Scenario from \cite{NHTSA}}
    \vspace*{1em}
    \label{appendix-2}
        \subsection{Intersection - Running Red Light}
            \includegraphics[width = 3.2in]{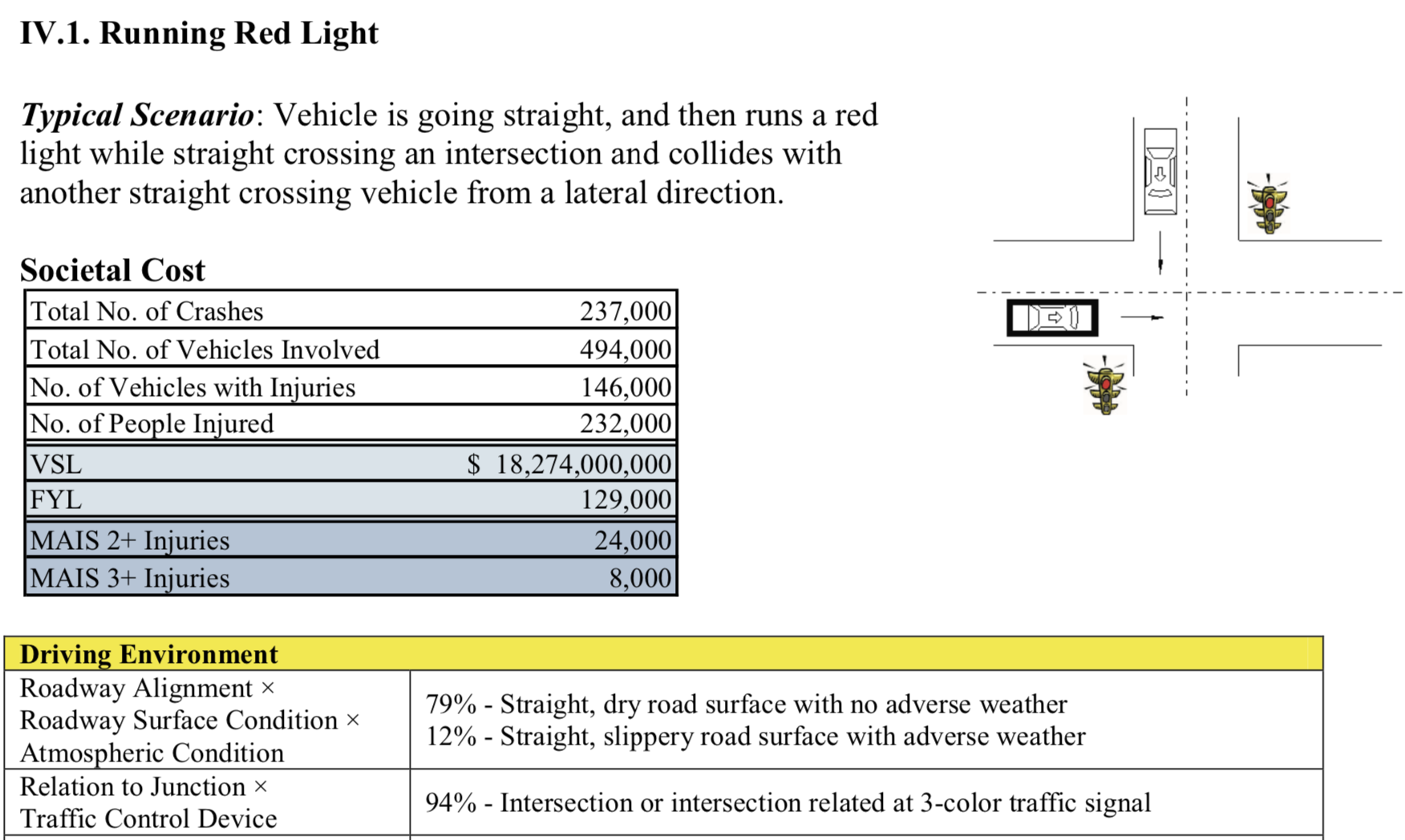}
            \vspace*{1em}
        \subsection{Non-intersection - Lane Change / Same Direction}
            \includegraphics[width = 3.2in]{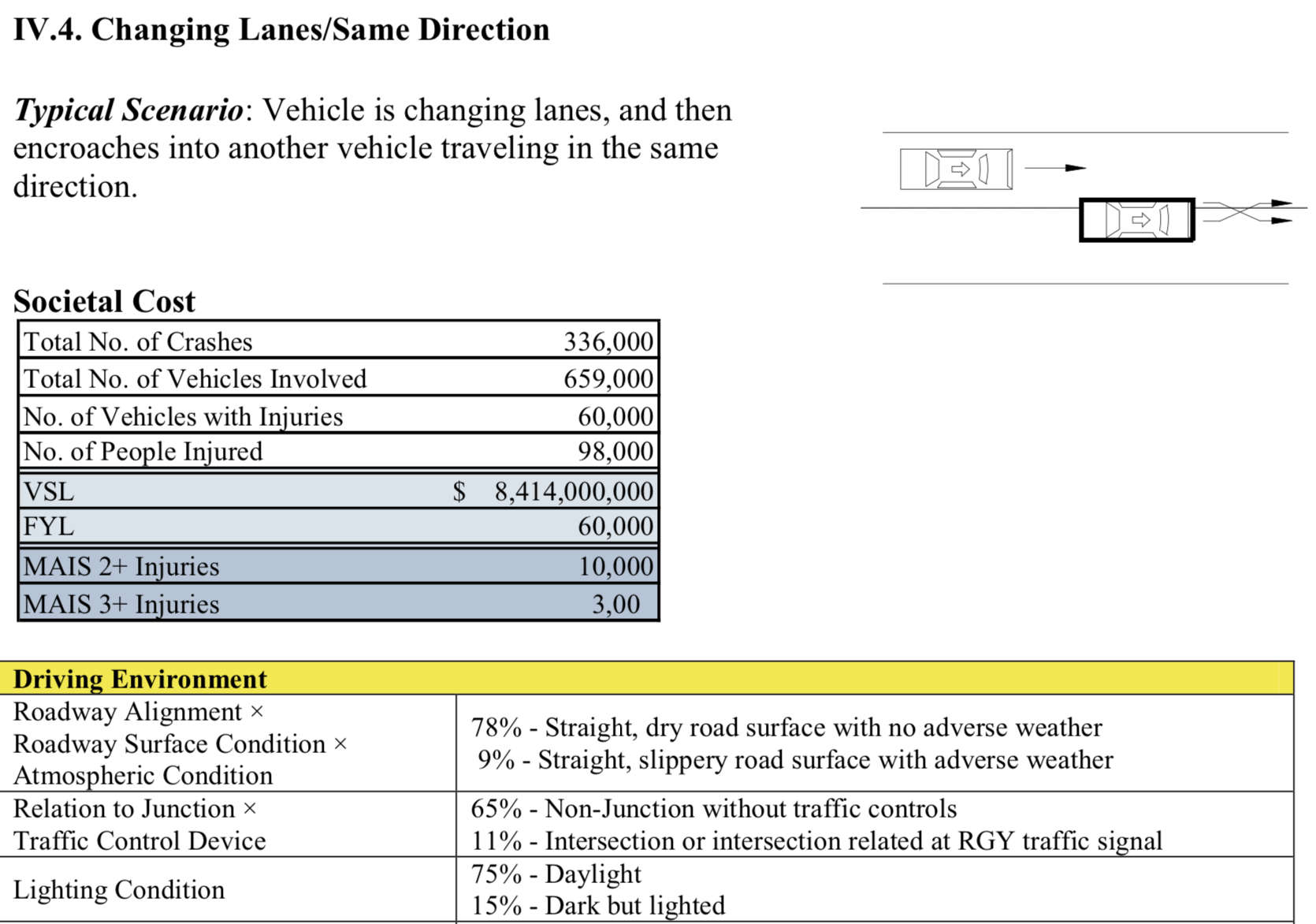}

\end{appendix}

\end{document}